\documentclass{article}

\usepackage{microtype}
\usepackage{graphicx}
\usepackage{epstopdf}
\usepackage{subfigure}
\usepackage{stfloats}

\usepackage{booktabs} 
\usepackage{graphicx} 
\usepackage{hyperref}
\usepackage{pifont}

\usepackage[accepted]{icml2024}

\usepackage{amsmath}
\usepackage{amssymb}
\usepackage{mathtools}
\usepackage{amsthm}
\usepackage{bm}
\usepackage{sidecap}

\newcommand{\mX}{\bm{X}}
\newcommand{\mY}{\bm{Y}}
\newcommand{\mZ}{\bm{Z}}

\newcommand{\mW}{\bm{W}}

\newcommand{\mA}{\bm{A}}
\newcommand{\mB}{\bm{B}}
\newcommand{\vx}{\bm{x}}
\newcommand{\vy}{\bm{y}}
\newcommand{\vz}{\bm{z}}
\newcommand{\vs}{\bm{s}}
\newcommand{\va}{\bm{a}}

\newcommand{\vu}{\bm{u}}

\newcommand{\vone}{\mathbf{1}}

\usepackage[capitalize,noabbrev]{cleveref}

\theoremstyle{plain}

\theoremstyle{definition}

\theoremstyle{remark}

\usepackage[textsize=tiny]{todonotes}

\icmltitlerunning{Fourier Controller Networks for Real-Time Decision-Making in Embodied Learning}

\begin{document}

\twocolumn[
\icmltitle{Fourier Controller Networks for Real-Time Decision-Making in \\ Embodied Learning}



\icmlsetsymbol{equal}{*}

\begin{icmlauthorlist}  
\icmlauthor{Hengkai Tan}{yyy}
\icmlauthor{Songming Liu}{yyy} 
\icmlauthor{Kai Ma}{yyy}
\icmlauthor{Chengyang Ying}{yyy}
\icmlauthor{Xingxing Zhang}{yyy}
\icmlauthor{Hang Su}{yyy}
\icmlauthor{Jun Zhu}{yyy}
\end{icmlauthorlist}

\icmlaffiliation{yyy}{Department of Computer Science and Technology, Institute for AI, Tsinghua-Bosch Joint ML Center, THBI Lab,
 BNRist Center, Tsinghua University, Beijing, 100084, China}

\icmlcorrespondingauthor{Hang Su}{suhangss@tsinghua.edu.cn}
\icmlcorrespondingauthor{Jun Zhu}{dcszj@tsinghua.edu.cn}


\icmlkeywords{Machine Learning, ICML}

\vskip 0.3in
]



\printAffiliationsAndNotice{}  

\begin{abstract}
Transformer has shown promise in reinforcement learning to model time-varying features for obtaining generalized low-level robot policies on diverse robotics datasets in embodied learning. 
However, it still suffers from the issues of low data efficiency and high inference latency. In this paper, we propose to investigate the task from a new perspective of the frequency domain. We first observe that the energy density in the frequency domain of a robot's trajectory is mainly concentrated in the low-frequency part. Then, we present the Fourier Controller Network (FCNet), a new network that uses Short-Time Fourier Transform (STFT) to extract and encode time-varying features through frequency domain interpolation.
In order to do real-time decision-making, we further adopt FFT and Sliding DFT methods in the model architecture to achieve parallel training and efficient recurrent inference.
Extensive results in both simulated (e.g., D4RL) and real-world environments (e.g., robot locomotion) demonstrate FCNet's substantial efficiency and effectiveness over existing methods such as Transformer, e.g., FCNet outperforms Transformer on multi-environmental robotics datasets of all types of sizes (from 1.9M to 120M). The project page and code can be found \href{https://thkkk.github.io/fcnet}{https://thkkk.github.io/fcnet}.

\end{abstract}


\section{Introduction}\label{sec:intro}

\begin{figure}[t]
\vskip 0.1in
\centering
\includegraphics[width=0.9\linewidth]{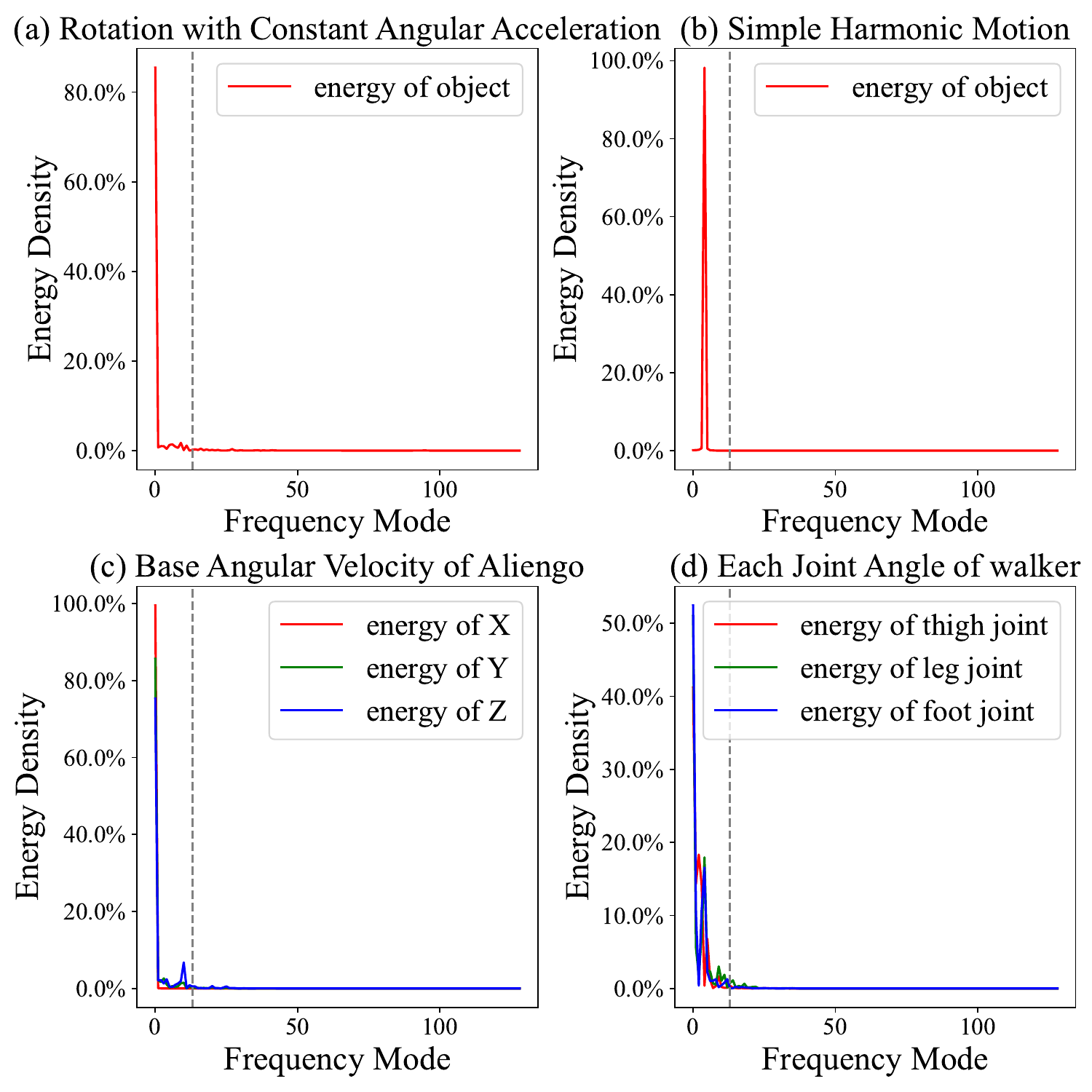}
\caption{Energy density (normalized to $0\%\sim100\%$ for all modes) in the frequency domain of different physical quantities across various motions. We choose $n = 256$ as the context length in the time domain. The time domain data represented in each of the four plots include: (a) Rotational motion with constant angular acceleration. (b) Simple harmonic motion. (c) Body angular velocity during a quadrupedal robot's run. (d) 
Joint angle of walker2d-expert-v2 in D4RL dataset.}
\label{fig:plot_energy_merged}
\vskip -0.1in
\end{figure}

Reinforcement Learning (RL) has been widely used in embodied learning scenarios~\cite{eth2020, rt2}, where agents interact with and learn from complex, dynamic physical environments.
Representative examples include robotic arm manipulation~\cite{gu2017deep, kalashnikov2018scalable, lee2022beyond}, legged robot locomotion in various challenging terrains~\cite{eth2020, eth2022, egocentric}, dexterous manipulation~\cite{gupta2016learning, rajeswaran2017learning, andrychowicz2020learning} and multi-task policies learning~\cite{kumar2022pre, kalashnikov2021mt, yu2020gradient}. Most of these RL methods use MLPs or RNNs to learn robotic control policies. 
Recently, a growing body of efforts has been devoted to pre-training on large-scale robotic datasets to obtain more generalized policies that can apply to various tasks~\cite{gato, vima, rt1, rt2, q_transformer, mobile_aloha}. 
Among various architectures, Transformer has been widely adopted because of its ability to model time-varying features and handle large-scale sequential robotics data.
In particular, DT~\cite{DT} is a pioneering and simple method that uses Transformer to directly model the trajectory sequence in RL auto-regressively, compared to traditional offline RL algorithms~\cite{kumar2020conservative} which use many mathematical tricks to learn from offline data.
The simplicity of the DT structure inspires follow-up work to use the attention mechanism to model the temporal features of diverse robotic trajectories. 
For instance, Gato~\cite{gato} serializes the Atari, Gym, chat, and image datasets (1.5T tokens) all into flat sequences for input to a Transformer, by utilizing its high capacity. 
RT-1~\cite{rt1} also utilizes Transformer for its multi-task integration capability for imitation learning on 130K expert trajectories of robotic control. 



\begin{table*}[t]
\vskip 0.1in
\caption{Comparision of Different Architectures. The energy density distribution of states in the frequency domain is often concentrated in the lowest $m$ modes, where $m\ll n$.
The term $b$ in the RetNet line represents the chunk size, which is set to $512$ in the paper\cite{retnet}.}
\begin{center}
\begin{small}
\begin{sc}
\begin{tabular}
{lp{2.6cm}p{2.6cm}p{2cm}p{3cm}p{2cm}}
\toprule
 &  Inference Cost & Training Cost  & Memory  & Parallel Training &  Performance \\ 
\midrule
MLP (history) & $\mathcal{O}(n)$ & $\mathcal{O}(n)$ & $\mathcal{O}(n)$ & \ding{52} &  \ding{56} \\
RNN & $\mathcal{O}(1)$ & $\mathcal{O}(n)$ & $\mathcal{O}(n)$ & \ding{56} &  \ding{52} \\
Transformer & $\mathcal{O}(n)$ & $\mathcal{O}(n^2)$ & $\mathcal{O}(n^2)$ & \ding{52} &  \ding{52} \\
RetNet & $\mathcal{O}(1)$ & $\mathcal{O}(nb)$ & $\mathcal{O}(n)$ & \ding{52} &  \ding{52} \\
FCNet & $\mathcal{O}(m)$ & $\mathcal{O}(mn (\log n + m))$ & $\mathcal{O}(nm)$ &\ding{52} &  \ding{52}\ding{52} \\
\bottomrule
\end{tabular}
\end{sc}
\end{small}
\end{center}
\vskip -0.1in
\label{tbl:comparision}
\end{table*}

One common feature of the existing works on Transformer is that they mainly focus on modeling the time-varying features of robotic trajectories in \emph{time domain} by drawing a direct analogy with that in modeling natural language sentences. 
We argue that this is insufficient for embodied learning, and instead we explore its unique features from the new angle of \emph{frequency domain}.
Specifically, we take a close examination on embodied learning in the {frequency domain} and observe that the energy density distribution of a robot's state sequence is mainly concentrated in the low-frequency part, as shown in Fig. \ref{fig:plot_energy_merged}. This is due to the inherent continuity and smoothness in natural physical phenomena and robot motor motion~\cite{kashiri2018overview}. 
However, existing works on Transformer and its variants directly model trajectories in the time domain, resulting in various issues. 
First, the data efficiency of existing Transformer architectures is low. They typically rely on large-scale data for good performance, while real-world physical data collection can be costly and time-consuming~\cite{rt1,rtx,mobile_aloha}. 
For instance, the dataset in RT-1~\cite{rt1} and subsequent work~\cite{rt2} requires 17 months to gather 130k episodes, out of reach for most research labs. 
Moreover, Transformer-based models cannot intrinsically reduce the computation complexity (e.g., $\mathcal{O}(n)$ for inference) and often struggle with the real-time processing requirements (i.e., low inference latency). 
For example, the inference frequency of such models is often around $3Hz$~\cite{rt1}. However, a typical  frequency of real-world legged robot control is above $50 Hz$~\cite{eth2020}.
Therefore, it is crucial to reduce the model's time complexity through algorithmic optimization, especially considering the constraints of robot hardware. 



\begin{figure}[h]
\vskip 0.1in
\centering
\includegraphics[width=\linewidth]{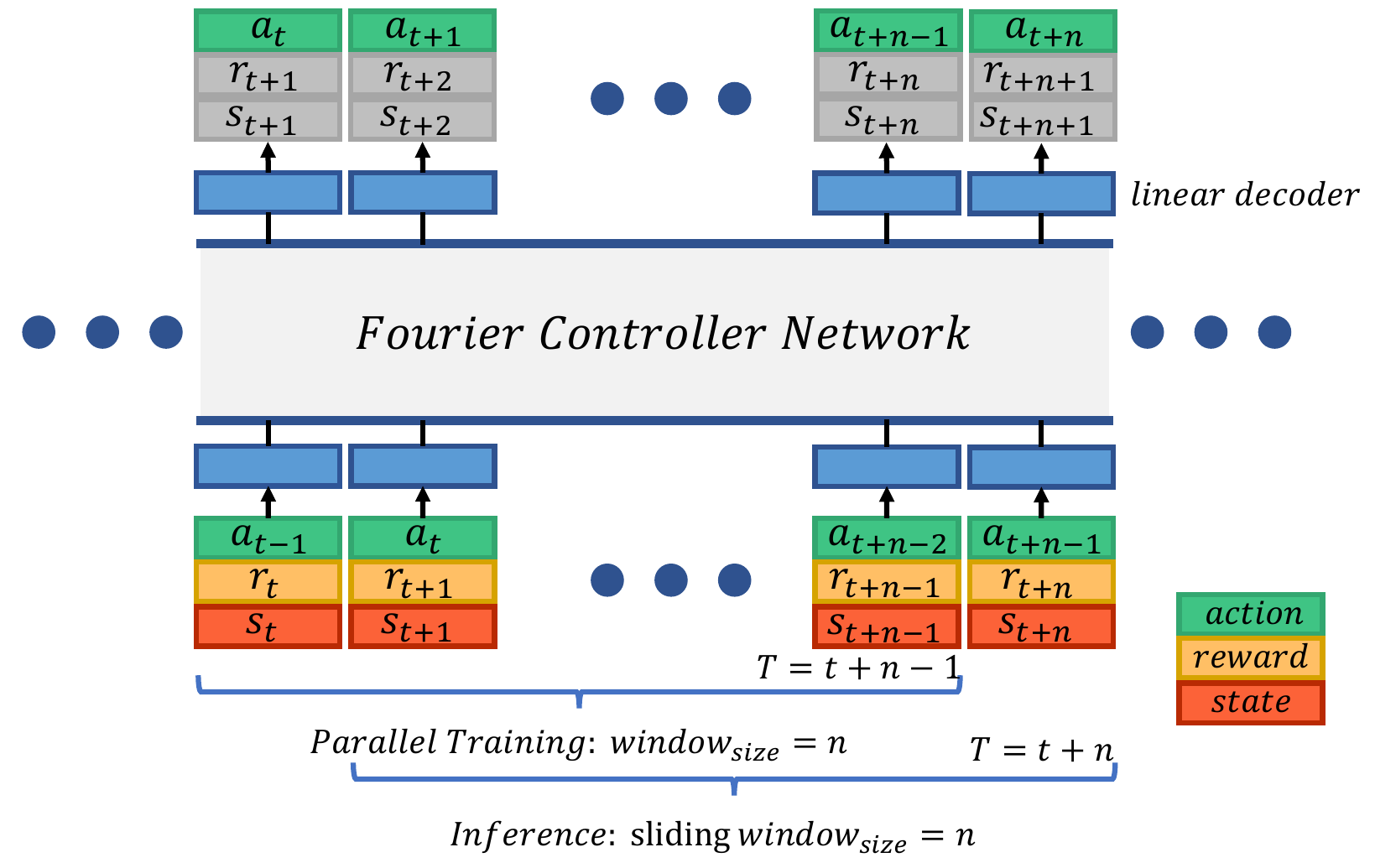}
\caption{Consider MDP as a sequence modeling problem. Especially in a continuous state space like embodied learning, it makes sense to perform Fourier-based modeling and training in the frequency domain of time-domain sequences within a window $[t, t+n-1]$. In the inference phase $T=t+n-1\rightarrow T=t+n$, the sliding window in Sec. \ref{sec:inference} is utilized for efficient inference. 
}
\label{fig:sequence_model}
\vskip -0.1in
\end{figure}

To address the aforementioned issues, we propose a new architecture of Fourier Controller Network (FCNet) based on the key observation in the frequency domain. FCNet grounds the inductive bias in robotic control inspired by the Fourier transform. 
We conceptualize low-level continuous control as a sequential decision-making problem. Our neural model is adept at predicting subsequent actions by analyzing a historical window of state data, as depicted in Fig.~\ref{fig:sequence_model}.
Guided by the observation in the frequency domain and the inductive reasoning that differential dynamics are simplified in the frequency domain (as suggested by~\cite{trefethen1996finite}), FCNet introduces a causal
spectral convolution (CSC) block.  It employs the Short-Time Fourier Transform (STFT) and linear transform for efficient feature extraction in the \emph{frequency domain}, distinct from Transformer and other prevalent architectures.
As shown in Fig.~\ref{fig:model_arch}, we focus on the $m$ lowest modes, with $m$ strategically selected to be $\ll n$, where $n$ is the length of the state window. 
Consequently, the high-frequency part in the frequency domain is filtered, allowing us to focus solely on these $m$ lowest modes. 
The CSC makes efficient training and inference possible, and has also been shown to have good performance in experiments. 

Furthermore, to achieve efficient parallel training and inference, 
which necessitates causality in the model's sequential outputs (dependent only on previous inputs) and the rapid generation of each output token for real-time response,
we introduce parallel training based on Fast Fourier transform (FFT), and recurrent inference based on sliding discrete Fourier transform (Sliding DFT) in FCNet.
As outlined in Table~\ref{tbl:comparision}, the FCNet demonstrates a computational complexity of $\mathcal{O}(mn \log n + m^2n)$ in parallel training setups, and $\mathcal{O}(m)$ for single-step inference. 
This efficiency marks a significant speed advantage over traditional Transformer models, enabling handling the complexities of real-time continuous control in dynamic environments.
We extensively evaluate FCNet in various settings. 
First, in the classic offline RL environments such as D4RL~\cite{d4rl}, FCNet outperforms Transformer, MLP-based methods as well as RetNet~\cite{retnet} which is a representative State Space Model (SSMs) and potential successor to Transformer. This shows strong feature extraction capabilities of FCNet from series of state data. 
Second, we evaluate Transformer and FCNet on a multi-environment robotics dataset. The results show that FCNet significantly outperforms Transformer with limited data, and also has lower inference latency and good scalability. We also verify the robustness of FCNet in real-world robots.
Finally, we test the inference latency of the Transformer (with KV cache) and FCNet under different hyperparameter settings related to model structure. The results show that the upward curve of the inference latency of FCNet is significantly slower than that of Transformer as the context length, number of layers, and hidden size are improved. This demonstrates the efficiency of the inference of FCNet.

\section{Background}

\subsection{Preliminary}
Real-world robotics control is often formulated as a Markov decision process (MDP) $(\mathcal{S}, \mathcal{A}, \mathcal{T}, r, \gamma)$, where $\mathcal{S} \subseteq \mathbb{R}^{d_s}$ and $\mathcal{A} \subseteq \mathbb{R}^{d_a}$ denote the state space and action space, respectively; $\mathcal{T}$ is the transition probability; $r$ is the reward function; and $\gamma \in (0,1)$ is a discount factor. The goal of robot control is to get a parameterized policy $\pi$ to take actions for interacting with the environment. 
Even in the fully-observed setting, the consideration of multi-task will introduce partial-observability~\cite{lee2020context,ghosh2021generalization,ying2023reward}, which demands encoding enough history information in the model. 
Following previous work~\cite{DT, egocentric}, we consider policies that map historical trajectories into action spaces $a_t = \pi(s_{\le t}, a_{< t})$ and maximize the accumulated expected rewards, i.e., 
\begin{equation}
    \mathbb{E}_{\tau \sim \pi_\theta} [R(\tau)]=\mathbb{E}_{\tau \sim \pi_\theta} \left[\sum_{i=0}^{\infty} \gamma^{i} r(s_i, a_i)\right],
\end{equation}
where $\tau$ represents the  trajectory of states and actions, defined as $\tau = \{s_0, a_0, s_1, a_1, \dots\}$. As acquiring the robotics dataset in the real world can be expensive, a well-used alternative method is to utilize expert datasets $D=\{\tau | \tau \sim \pi_\beta, \tau=(s_0, a_0, \dots, s_T, a_T)\}$ for imitation learning, or offline datasets $D=\{\tau | \tau \sim \pi_\beta, \tau=(s_0, a_0, r_0, \dots, s_T, a_T, r_T)\}$ for offline RL. Here $\pi_{\beta}$ is an unknown behavior policy.  We aim to train a policy $\pi_\theta$ using the dataset, with the goal of maximizing the expected return $\mathbb{E}_{\tau \sim \pi_\theta} [R(\tau)]$.

For expert datasets that do not include reward information, our objective is to ensure that the behavior of the trained policy $\pi_{\theta}$ closely aligns with the expert behavior policy $\pi_{\beta}$. Thus, the optimization objective can be summarized as
\begin{equation}
    \mathcal{L}(\theta) =  {1 \over T|D|} \sum_{\tau \in D} \sum_{i=0}^T d(a_i, \hat a_i),
\label{eq:imitation}
\end{equation}
where $\hat{a}_i$ is the action derived from $\pi_{\theta}$, $a_i$ is the action in $\tau$, and $d$ is the distance function. 
Typically, when $\pi_{\theta}$ takes into account historical trajectories, it employs a context window of length $n$ representing the extent of historical dependencies, i.e., $a_t = \pi_\theta(s_{t-n+1:t}, a_{t-n+1:t-1})$. In practice, we can choose $d$ in Eq.~(\ref{eq:imitation}) as the mean-squared error (MSE) for deterministic actions, and the Kullback-Leibler (KL) Divergence for stochastic actions. 

For offline datasets that include reward information, one method is to incorporate the return-to-go, i.e., the suffix of the reward sequence, into the policy's sequence input. This often results in a policy with higher performance than the original behavior policy in the dataset, as the results of DT~\cite{DT} show.


\paragraph{Time Complexity of Transformer} 
In addition to the DT just mentioned, there is a lot of work on pre-training robotics datasets using Transformer \cite{gato, vima, rt1, rt2, q_transformer}, and they all use trajectories analogous to language as input to  Transformer.
The training time complexity of the Transformer model is $\mathcal{O}(n^2d_h+nd_h^2)$, where $n$ is the length of both the input and output sequences and $d_h$ is the hidden dimension.  The inference time complexity of the Transformer with KV cache is $\mathcal{O}(nd_h+d_h^2)$. 
In scenarios where $d_h$ is relatively stable,
we primarily consider the impact of $n$ on time complexity. Consequently, the training and inference time complexities of the Transformer can be simplified to $\mathcal{O}(n^2)$ and $\mathcal{O}(n)$, respectively.


\subsection{Motivation}
\label{sec:motivation}

As previously mentioned, existing architectures such as Transformer mainly focus on modeling the time-varying features of robotic trajectories in the time domain, through mechanisms such as attention. However, they do not take into account the special properties embedded in robotic control in the frequency domain. Below, we start with simple physical motions to illustrate the significance of modeling temporal features in the frequency domain.

Consider the motion of a mass along the x-axis as a simple example. We assume knowledge of the mass's state at $n$ distinct moments. The complexity of describing its motion depends on the nature of the movement. 
For a particle at rest, only the initial position $x_0$
is needed. In uniform linear motion, both the initial position $x_0$ and velocity $v$ are required. For uniformly accelerated linear motion, three quantities are essential: initial position $x_0$, initial velocity $v$, and acceleration $a$, resulting in the equation $x=x_0+v_0 t + at^2/2$. Similarly, simple harmonic motion, which is common in nature, requires amplitude $A$, angular frequency $\omega$, and phase $\phi$ for the equation $x=A\sin {(\omega t + \phi)}$.
In fact,  real motor movements are typically smooth to minimize energy losses and to facilitate their application in robotic arms~\cite{kashiri2018overview}. For instance, uniformly accelerated motion is employed in the trapezoidal acceleration and deceleration curves of motors, and simple harmonic motion (or sinusoidal function) is used in the S-curves of motors, both common in motor motion.

This insight leads us to recognize that natural object motion often adheres to physical principles, like energy conservation, and optimal trajectories are typically smooth, possibly with minor irregularities. Viewing this from a data-driven standpoint, if we have details on a mass's position, velocity, and other states at $n$ moments, the quantity of information embedded in this does not need to be described using  $\mathcal{O}(n)$ level parameters. In fact, a significantly smaller set of parameters might be adequate for accurate representation.
Similarly, complex rigid structures such as robots can also be approximated in this manner, given that robots are composed of rigid components like joints and links. Typically, a robot's state includes various parameters such as motor position, motor velocity, and body angular velocity, among others. 
The flexible motion of the robot can be characterized by a limited set of physical quantities. 
However, the measured state variables often contain a significant amount of high-frequency noise, due to sensor uncertainties. 
Therefore, we propose to transform the state representation of the object on $n$ moments into the frequency domain, rather than modeling the trajectories in the time domain. 
By filtering out high-frequency components and retaining low-frequency ones, we can more accurately capture the essential motion information of the object.


Indeed, as demonstrated in Fig. \ref{fig:plot_energy_merged}, both the common mechanical motions and the motions of real robots show a consistent pattern in the frequency domain, namely, 
their energy density distributions predominantly reside in a low-frequency region. This observation aligns with our hypothesis. Additionally, our empirical analyses reveal that in various scenarios, the number $m$ of significant low-frequency modes for robot states is much smaller than $n$. We empirically set the value 
$m \ll n$
in the experiment. Consequently, this allows us to model state sequences over time with reduced complexity compared to the Transformer, thanks to frequency domain interpolation. 

\section{Fourier Controllers}
We now formally introduce Fourier Controller Network (FCNet) for efficient robotic control. FCNet incorporates an inductive bias about frequency domain features and aims to minimize the time complexity of both training and inference.

\begin{figure*}[th]
\vskip 0.1in
\centering
\includegraphics[width=0.9\linewidth]{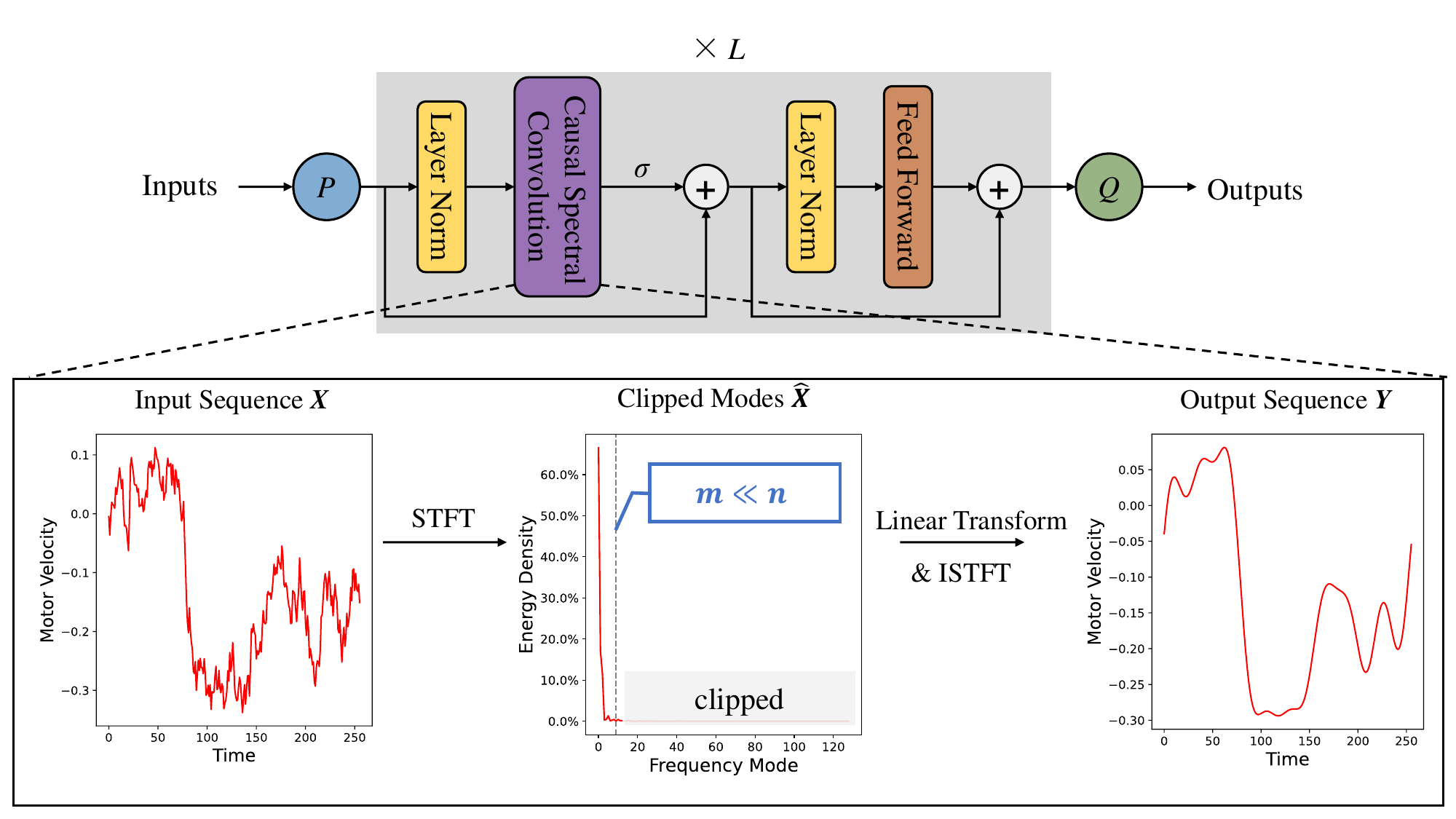}
\caption{The overall model architecture of FCNet. In order to ensure efficient training and inference, we cannot apply Fourier transform to all history trajectories, but instead apply STFT to a window of historical data of length $n$ to filter high-frequency part in the frequency domain, and then apply linear transform and inverse STFT back to the time domain. That is, the CSC block comes in the frequency domain to model temporal features, while the FFN is used to model features on the hidden dimension. $P$ is point-wise encoder and $Q$ is point-wise decoder mentioned in 
Sec. \ref{sec:overall_arch}. $\sigma$ is the activation function.}
\label{fig:model_arch}
\vskip -0.1in
\end{figure*}


\subsection{Overall Architecture}
\label{sec:overall_arch}
As shown in Fig.~\ref{fig:model_arch}, FCNet comprises a position-wise encoder $P\colon \mathbb{R}^{d_s} \rightarrow \mathbb{R}^{d_h}$, $L$ identical stacked Fourier layers, and a position-wise decoder $Q\colon \mathbb{R}^{d_h} \rightarrow \mathbb{R}^{d_a}$, where both $P$ and $Q$ are parameterized by feed-forward networks (FFNs) and $d_h$ denotes the hidden dimension. Each Fourier layer consists of two primary modules: a causal spectral convolution (CSC) block and a position-wise FFN block. Given an input window of historical states, denoted as  $\mX^0:= [\vx_0^0, \dots, \vx^0_{n-1}]^\top \in \mathbb{R}^{n\times d_s}$ where $ \vx^0_i:=  \vs_i $ for $ \forall 0 \le i < n$, our model outputs the predicted actions for corresponding time steps: ${\mX}^{L+2}:= [{\vx}^{L+2}_0, \dots, {\vx}^{L+2}_{n-1}]^\top \in \mathbb{R}^{n\times d_a}$ with $ {\vx}^{L+2}_i:=  (\hat \va_i) $ for $ \forall 0 \le i < n$.
The computation process of the FCNet is:
\begin{equation}
\begin{aligned}
    \mX^1\in \mathbb{R}^{n\times d_h} &= P(\mX^0),\\
    \mY^l &= \mathrm{gelu}(\mathrm{CSC}(\mathrm{LN}(\mX^l))) + \mX^l,\\
    \mX^{l+1} &= \mathrm{FFN}(\mathrm{LN}(\mY^l)) + \mY^l,\\
    \mX^{L+2} \in \mathbb{R}^{n\times d_a} &= Q(\mX^L),\\
\end{aligned}
\end{equation}
where $l \in \{1,\dots,L\}$, $\mathrm{LN}(\cdot)$ is the LayerNorm~\cite{ba2016layernorm}, and $P$ is a single-layer FFN, while $\mathrm{FFN}(\cdot) $ and $Q$ are two-layer FFNs. $\mathrm{gelu}(\cdot)$ refers to the Gaussian Error Linear Unit (GELU) activation~\cite{gelu}. Of note, the parameters for each layer's $\mathrm{FFN}(\cdot)$ are not shared.

In order to leverage good scalability of sequence modeling~\cite{gpt3}, we can supervise the action of the policy output $X^{L+2}$ in a supervised learning manner\cite{DT}. Based on Eq. (\ref{eq:imitation}), the training objective can be formulated as:
\begin{equation}
    \mathcal{L}(\theta) =  {1 \over |D|} \sum_{\tau \in D}  d(\va_{0:n}, X^{L+2}).
\end{equation}



\subsection{Causal Spectral Convolution}
In order to model time-varying features in the frequency domain, in this subsection, we delve into the mechanism of causal spectral convolution (CSC) block, which is a trainable block and utilizes the Short-Time Fourier Transform (STFT) to efficiently extract and encode time-varying features. It is nontrivial to encode because it is necessary to ensure that the output is causal. 
Given an input sequence $\mX:= [\vx_0, \dots, \vx_{n-1}]^\top \in \mathbb{R}^{n\times d_h}$, the output of the CSC $\vy_{n-1} \in \mathbb{R}^{d_h}$ can be formalized as:
\begin{equation}
\label{eq:csc}
\begin{aligned}
    \hat{\vx}_k &= \mathcal{F}(\mX)(k), &&\hat{\vx}_k \in \mathbb{C}^{d_h}, 0\le k < m, \\ 
    \hat{\mY} &= \mW  \hat{\mX}  ,&& \hat{\mX}  = [\hat{\vx}_0, \dots, \hat{\vx}_{m-1}]^\top,\\
    & && \hat{\mY}  = [\hat{\vy}_0, \dots, \hat{\vy}_{m-1}]^\top,\\
    \vy_{n-1} &= \mathcal{F}^{-1}(\hat{\mZ})(n-1),&& \hat{\mZ} = [\hat{\vz}_0, \dots, \hat{\vz}_{n-1}]^\top,
\end{aligned}
\end{equation}
where $\hat{\mX}, \hat{\mY}\in \mathbb{C}^{m \times d_h}$ and $ \hat{\mZ} \in \mathbb{C}^{n \times d_h}$. $1 \le m \le 1+\lfloor n/2 \rfloor$ is a hyper-parameter denoting the number of modes preserved in the frequency domain, $\mW \in \mathbb{C}^{m \times m}$ is the weight matrix, and $\mZ$ is obtained by extending $\mY$ according to the conjugate symmetry of real series:
\begin{equation}
    \hat{\vz}_k = \begin{cases}
        \begin{aligned}
        &\hat{\vy}_k, &&0 \le k < m,\\
        &\overline{\hat{\vy}_{n-k}},  &&n-m+1\le k<n,\\
        &0, &&\text{otherwise}.
        \end{aligned}
        \end{cases}
\label{eq:z_y}
\end{equation}
Here, $\mathcal{F}$ and $\mathcal{F}^{-1}$ denote the Discrete Fourier Transform (DFT) and its inverse (IDFT):
\begin{equation}
\begin{aligned}
    \mathcal{F}(\mX)(k) &= \sum_{i=0}^{n-1} \vx_i \exp{\left( - \frac{j2\pi}{n}ki \right)}, \\
    \mathcal{F}^{-1}(\hat{\mZ})(i) &= \frac{1}{n}\sum_{k=0}^{n-1} \hat{\vz}_k \exp{\left(  \frac{j2\pi}{n}ki \right)},
\end{aligned}
\label{eq:dftidft}
\end{equation}
where $j=\sqrt{-1}$ is the imaginary unit.

In Eq.~\eqref{eq:csc}, we start by encoding the input sequence via the DFT, where only low-frequency components are preserved while high-frequency noises are filtered by using only $m$ DFTs, according to the smoothness of the physical states. 
Then, a linear transform is applied in the frequency domain, followed by an IDFT back to the physical domain. It is noted that the output $\vy_{n-1}$ only depends on the history $\vx_{i},i\le n-1$, ensuring the causality of the prediction.

\subsection{Parallel Training}
\label{section:parallel_training}
Parallelization significantly speeds up model training by distributing tasks across multiple processors, especially for large datasets.
Here we fully parallelize the FCNet training to improve efficiency.
The FFN can be easily parallelized by matrix partitioning, and thus we mainly discuss the parallelization of the CSC block (see Appendix~\ref{sec:appA} for its schematic diagram).
The CSC block involves operations that are inherently sequential, and it is nontrivial to parallelize due to dependencies between operations.

\paragraph{Representation of computation process.} 
Adopting this approach, for a given input sequence $\mX:= [\vx_0, \dots, \vx_{n-1}]^\top \in \mathbb{R}^{n\times d_h}$, the CSC block outputs in parallel $\mY:= [\vy_0, \dots, \vy_{n-1}]^\top \in \mathbb{R}^{n\times d_h}$, where $\vy_k$ is solely dependent on $\vx_i$ for $i \le k$. 
Our goal is to compute each $\vy_t$ ($0 \le t < n$) quickly and in parallel. Consider that we are computing the output $\vy_t$ ($0 \le t < n$) at moment t. $\vy_t$ can only depend on $\vx_i$ ($i \le t$) for causality. Therefore, the $\hat \mX$ in the frequency domain for each $\vy_t$ is distinct. 
We use $\hat \mX^{(t)}$ to denote the $\hat \mX$ corresponding to $\vy_t$. From Eq.~(\ref{eq:csc}) and Eq.~(\ref{eq:dftidft}), we have:
\begin{equation}
\begin{aligned}
    \hat \vx^{(t)}_k &= \sum_{i=t-n+1}^{t} \vx_i \exp{\left( - \frac{j2\pi}{n}ki \right)},
\end{aligned}
\label{eq:direct_training}
\end{equation}

To simplify the expression, we let $u_k = \exp{(\frac{j2\pi}{n}k )}$ and $\vu = [u_0, \dots,u_{m-1}]^\top \in \mathbb{C}^m$. Each $\hat \vx^{(t)}_k$ can be expressed in terms of its previous $\hat \vx_k$:
\begin{equation}
\begin{aligned}
    \hat \vx^{(t)}_k &= u_k(\hat \vx^{(t-1)}_k + x_t - x_{t-n}) 
    = u_k(\hat \vx^{(t-1)}_k + f_t), \\
    \Rightarrow \hat \mX^{(t)} &= \vu \odot (\hat \mX^{(t-1)} + \vone_m f_t^\top), \label{eq:recurrent_training}
\end{aligned}
\end{equation}
where $f_t = x_t - x_{t-n} \in \mathbb{R}^{d_h}$ can be computed in parallel, $
    \hat{\mX}^{(t)}= [\hat{\vx}^{(t)}_0, \dots, \hat{\vx}^{(t)}_{m-1}]^\top \in \mathbb{C}^{m \times d_h}$
and $\odot$ means Hadamard product or element-wise product. The above equation can be computed recursively for each $\hat \mX^{(t)}$. 

After determining $\hat \mX^{(t)}$ for $ 0 \le t < n$, we can obtain $\hat{\mY}^{(t)}, \hat{\mZ}^{(t)}, y_t, 0 \le t < n$ using Eq.~(\ref{eq:csc}) and Eq.~(\ref{eq:z_y}).

\paragraph{Parallel training.} 
The representation in Eq.~\eqref{eq:direct_training} has time complexity $\mathcal{O}(md_h n^2)$ in parallel and Eq.~\eqref{eq:recurrent_training} has a \textbf{serial} $\mathcal{O}(md_h n)$ time complexity.
Here, we seek to find a training process that can be parallelized with low time complexity.
Let us dissect the recursive formula given in Eq.~\eqref{eq:recurrent_training} for $\hat \mX^{(t)}$ where $ 0 \le t < n$.
We have
\begin{equation}
\begin{aligned}
    \hat \mX^{(t)} &= \vu^{\circ t} \odot \hat \mX^{(0)} + \sum_{i=0}^{t-1} \vu^{\circ (i+1)} f_{t-i}^\top,
\end{aligned}
\label{eq:recur_expand}
\end{equation}

where $\vu^{\circ t}$ means $\underbrace{ \vu \odot \vu \odot \dots \odot \vu}_{\text{$t$~elements}}$. Let
\begin{equation}
\begin{aligned}
    \mA_i &= [(\vu^{\circ (i+1)})_{\times d_h}] \in \mathbb{C}^{m\times d_h}, 0\le i < n-1, \\
    \mB_i &= [(f_i^\top)_{\times m}] \in \mathbb{C}^{m\times d_h}, 1 \le i < n,
\end{aligned}
\end{equation}
and for special case $\mA_{n-1}=\mA_{n-2}$, $\mB_0 = \hat X^{(0)} \in \mathbb{C}^{m\times d_h}$. From Eq.~\eqref{eq:recur_expand} we get:
\begin{equation}
\begin{aligned}
    \hat \mX^{(t)} &= \sum_{i=0}^t A_{i} \odot B_{t-i}, &&0 \le t < n  ,
\end{aligned}
\end{equation}
which is a typical form of FFT convolution $\{\hat \mX^{(t)}\} = \{A_{i}\} * \{B_{i}\}$ where $*$ represents linear convolution. We can compute all $\hat \mX^{(t)}, 0 \le t < n $ by FFT with time complexity $\mathcal{O}(md_h n\log n )$ \textbf{in parallel}. The fast parallel computation of the IDFT  is homogeneous. Thus the training time complexity of the whole CSC block is $\mathcal{O}(md_hn\log n + m^2d_hn)$ because of the computation of $\hat{\mY}^{(t)} = \mW  \hat{\mX}^{(t)}, 0 \le t < n$. Therefore, the total training time complexity is $\mathcal{O}(md_hn \log n + m^2d_hn + nd_h^2)$ or reduced to $\mathcal{O}(mn \log n + m^2n)$.

Furthermore, FCNet exhibits a memory complexity of $\mathcal{O}(mnd_h)$ (can be reduced to $\mathcal{O}(mn)$) during training due to the storage for  $\hat x^{(t)}_k$ (where $0\le t<n$ and $0\le k < m$). 

\subsection{Reprensentation of Recurrent Inference} 
\label{sec:inference}

Efficient real-time inference is important for the deployment of robots~\cite{sandha2021sim2real}. It is difficult to utilize time-domain information to achieve efficient inference. Therefore we utilize the cache of frequency domain information for fast inference. 
During inference, supposing that we cache the results computed in the last time step, the CSC can recurrently compute the current output $\vy_{n-1}$ given a newly coming $\vx_{n-1}$ using Sliding DFT (as well as Eq. (\ref{eq:recurrent_training})):
\begin{equation}
    \hat{\vx}_k = \exp{\left(  \frac{j2\pi}{n}k \right)} (\hat{\vx}_k' - \vx_{-1} + \vx_{n-1}),
\label{eq:inference}
\end{equation}
where $0\le k < m$. After obtaining $\hat{\vx}_0, \dots, \hat{\vx}_{m-1}$, we compute $\vy_{n-1}$ as in Eq.~(\ref{eq:csc}). Here, $\hat{\vx}_k'$ is the result cached in the last time step, which is equal to $\mathcal{F}(\mX')(k), \mX':= [\vx_{-1}, \dots, \vx_{n-2}]^\top$. Because we cached the $\hat{\vx}_k'$ in the last time step, the time complexity of Eq. (\ref{eq:inference}) is $\mathcal{O}(m d_h)$ instead of $\mathcal{O}(m d_h n)$. 

After computing $\hat{\mX} = [\hat{\vx}_0, \dots, \hat{\vx}_{m-1}]^\top$, the $\hat{\mY} = \mW  \hat{\mX}$ and $\mathcal{F}^{-1}(\hat{\mZ})(n-1)$ part in Eq. (\ref{eq:csc}) can be preprocessed by calculating the result of multiplying $\mW$ by the corresponding vector of $\mathcal{F}^{-1}(\hat{\mZ})(n-1)$ in advance. Therefore, the total inference time complexity is $\mathcal{O}(md_h + d_h^2)$ (where $\mathcal{O}(d_h^2)$ corresponds to FFN), further reduced to $\mathcal{O}(m)$. 
This mechanism is advantageous for rapid inference since it allows for the generation of actions sequentially during the evaluation of an embodied agent.





\section{Experiments}
In this section, we conduct extensive experiments to evaluate the performance and efficiency of FCNet.  
Initially, FCNet's effectiveness in fine continuous control is compared against Transformer and RetNet (a current SSM architecture) using the D4RL offline RL benchmark. 
Next, FCNet and Transformer are assessed on a multi-environment legged robot locomotion dataset via imitation learning, highlighting FCNet's adaptability across various environments, particularly with scarce data.
Finally, we investigate the inference efficiency of FCNet compared to Transformer under diverse model structure hyperparameter settings. 

\subsection{Evaluation in Offline RL}
\label{sec:d4rl}
To evaluate FCNet in the D4RL~\cite{d4rl} offline RL tasks, we employ a broad range of baselines featuring various network architectures such as MLP, Transformer, and RetNet. 
We start with classic MLP-based methods like Behavior Cloning (BC), CQL~\cite{kumar2020conservative}, BEAR~\cite{kumar2019stabilizing}, BRAC-v~\cite{wu2019behavior}, and AWR~\cite{peng2019advantage}. Additionally, we incorporate the transformer-based method DT~\cite{DT} and the RetNet-based method DT-RetNet, which replaces the Transformer architecture in DT into RetNet.
In D4RL evaluations, FCNet follows some design strategies of DT, i.e., concatenating action, return-to-go (normalized, with $1.0$ indicating an expert policy), and state into a single token $(a^\top, R^\top, s^\top)^\top$, and then predicts the next token autoregressively. As shown in Table~\ref{tbl:mujoco_results}, our FCNet is competitive with DT, and significantly surpasses other methods. 
Particularly, FCNet outperforms DT-RetNet, suggesting that FCNet's inductive bias for robotic control has more powerful feature extraction capability and performance compared to SSMs. 

In addition to locomotion tasks, we expand our investigation to encompass manipulation tasks, specifically the Adroit suite in the D4RL dataset, which comprises various robotic arm manipulation challenges and are not harmonic, distinct from locomotion tasks. 
Our results, as detailed in the Table~\ref{tbl:adroit_results}, demonstrate the performance of FCNet in comparison with DT, BC, and CQL. 
FCNet beats DT on 9 of the 12 manipulation tasks in D4RL-Adroit. These results highlight FCNet's robust performance across a range of tasks, outperforming DT, BC, and CQL in the domain of manipulation. This demonstrates the broad applicability of our approach beyond locomotion tasks.

\begin{table*}[th]
\caption{
\textbf{Results in D4RL-Mujoco}.
For all algorithms, we report the performance with mean and variance under three random seeds.
} 
\vskip 0.1in
\centering
\small
\resizebox{\textwidth}{!}{
\begin{sc}
\begin{tabular}{lrrrrrrrrr}
\toprule
\multicolumn{1}{c}{\bf{Task Name}} & \multicolumn{1}{c}{\bf FCNet(Ours) } & \multicolumn{1}{c}{\bf DT } & \multicolumn{1}{c}{\bf DT-RetNet } & \multicolumn{1}{c}{\bf BC} & \multicolumn{1}{c}{\bf CQL} & \multicolumn{1}{c}{\bf BEAR} & \multicolumn{1}{c}{\bf BRAC-v} & \multicolumn{1}{c}{\bf AWR}\\
\midrule
halfcheetah-medium-expert & $\bf{91.2\pm 0.3}$ &  $86.8 \pm 1.3$ & $\bf{91.4\pm 1.2}$ & $35.8$  &  $62.4$ &  $53.4$ &  $41.9$ & $52.7$ \\
walker2d-medium-expert  & $\bf{108.8\pm 0.1}$ &  $\bf{108.1} \pm 0.2$ & $79.1\pm1.8$ & $6.4$ &  $98.7$ &  $40.1$ &  $81.6$ & $53.8$  \\
hopper-medium-expert      & $\bf{110.5\pm 0.5}$ &  $107.6 \pm 1.8$  & $102.9\pm5.6$ & $\bf{111.9}$ & $\bf{111.0}$ &  $96.3$ &   $0.8$ & $27.1$\\
\midrule
halfcheetah-medium & $42.9\pm 0.4$  &  $42.6 \pm 0.1$  & $43.2\pm 0.4$ & $36.1$     &  $44.4$    &  $41.7$ &  $\bf{46.3}$ & $37.4$ \\
walker2d-medium      &  $75.2\pm 0.5$  &  $74.0 \pm 1.4$ & $73.3\pm 4.3$ & $6.6$    &  $\bf{79.2}$ &  $59.1$  & $\bf{81.1}$ & $17.4$ \\
hopper-medium      &  $57.8\pm 6.0$ &  $67.6 \pm 1.0$  & $\bf{74.1}\pm 5.3$ & $29.0$  &  $58.0$ &  $52.1$ &  $31.1$ & $35.9$\\
\midrule
halfcheetah-medium-replay &  $39.8\pm 0.8$ &  $36.6 \pm 0.8$  & $17.2\pm 4.3$ & $38.4$  &  $46.2$ &  $38.6$ &  $\bf{47.7}$ & $40.3$\\
walker2d-medium-replay &  $63.5\pm 7.5$ &  $\bf{66.6} \pm 3.0$& $32.4\pm6.6$ & $11.3$  &  $26.7$ &  $19.2$ &  $0.9$ & $15.5$\\
hopper-medium-replay      &  $\bf{85.8\pm 1.7}$ &  $82.7 \pm 7.0$  & $58.4\pm 2.8$ & $11.8$ &  $48.6$ &  $33.7$ &   $0.6$ & $28.4$\\
\midrule
\multicolumn{1}{c}{\bf Average} &  $\bf{75.1}$ &  $\bf{74.7}$ & $63.6$ & $46.4$ & $63.9$ & $48.2$ & $36.9$ & $34.3$ \\
\bottomrule
\end{tabular}
\end{sc}
}
\label{tbl:mujoco_results}
\vskip -0.1in
\end{table*}

\begin{table*}[th]
\caption{
\textbf{Results in D4RL-Adroit}.
For FCNet and DT, we report the performance with mean and variance under five random seeds. Other scores are reported by the D4RL paper. 
} 
\vskip 0.1in
\centering
\small
\begin{sc}
\begin{tabular}{lrrrr}
\toprule
\multicolumn{1}{c}{} & \multicolumn{1}{c}{\bf FCNet (Ours)} & \multicolumn{1}{c}{\bf DT} & \multicolumn{1}{c}{\bf BC} & \multicolumn{1}{c}{\bf CQL} \\
\midrule
pen-human & $57.7\pm11.1$ & $-0.2\pm1.8$ & 34.4 & 37.5 \\
hammer-human & $1.2\pm0.0$ & $0.3\pm0.0$ & 1.5 & 4.4 \\
door-human & $0.4\pm0.5$ & $0.1\pm0.0$ & 0.5 & 9.9 \\
relocate-human & $0.2\pm0.2$ & $0.0\pm0.0$ & 0.0 & 0.2 \\
pen-cloned & $50.4\pm24.1$ & $22.7\pm17.1$ & 56.9 & 39.2 \\
hammer-cloned & $0.2\pm0.0$ & $0.3\pm0.0$ & 0.8 & 2.1 \\
door-cloned & $-0.2\pm0.0$ & $0.1\pm0.0$ & -0.1 & 0.4 \\
relocate-cloned & $-0.2\pm0.0$ & $-0.3\pm0.0$ & -0.1 & -0.1 \\
pen-expert & $108.0\pm11.3$ & $110.4\pm20.9$ & 85.1 & 107.0 \\
hammer-expert & $121.1\pm6.1$ & $89.7\pm24.6$ & 125.6 & 86.7 \\
door-expert & $102.9\pm2.9$ & $95.5\pm5.7$ & 34.9 & 101.5 \\
relocate-expert & $50.0\pm6.0$ & $15.3\pm3.6$ & 101.3 & 95.0 \\
\midrule
\multicolumn{1}{c}{\bf Average} & \bf{41.0} & 27.8 & 36.7 & 40.3 \\
\bottomrule
\end{tabular}
\end{sc}
\label{tbl:adroit_results}
\vskip -0.1in
\end{table*}

\subsection{Evaluation in Legged Robot Locomotion}
\label{sec:legged_exp}

To test FCNet on the more challenging multi-environment legged robot locomotion, we introduce a new, substantially larger dataset than that in D4RL. Compared with Transformer-based methods in this dataset, our focus is to showcase FCNet's strong fitting ability even with limited data, and further real-time inference on a real-world robot. 

\paragraph{Multi-Environment Legged Robotics Dataset.} 
\label{para:dataset}
In order to evaluate the performance of the policy in modeling historical information sequences in a complex multi-task environment, which is rarely available in previous d4rl datasets, We develop a multi-environment legged robot locomotion dataset, which covers various skills (e.g., standing, rushing, crawling, squeezing, tilting, and running), and different terrains (e.g., rough terrain, stairs, slopes, and obstacles), following~\citet{eth2022}. The dataset is collected in Isaacgym~\cite{isaacgym} by some expert-performance RL policies which have demonstrated strong performance in the real world. 
It is noteworthy that this dataset is designed for practical robotics applications. Models trained using this dataset can be directly deployed on real-world legged robots (e.g., Unitree Aliengo)
to perform a variety of locomotive skills across diverse terrains. 
The test version of the dataset contains the aforementioned skills and terrains, with 320,000 trajectories and 60M steps.\footnote{We make some of the data public on the project page.}

\begin{figure}[t]
\vskip 0.1in
\centering
\includegraphics[width=0.9\linewidth]{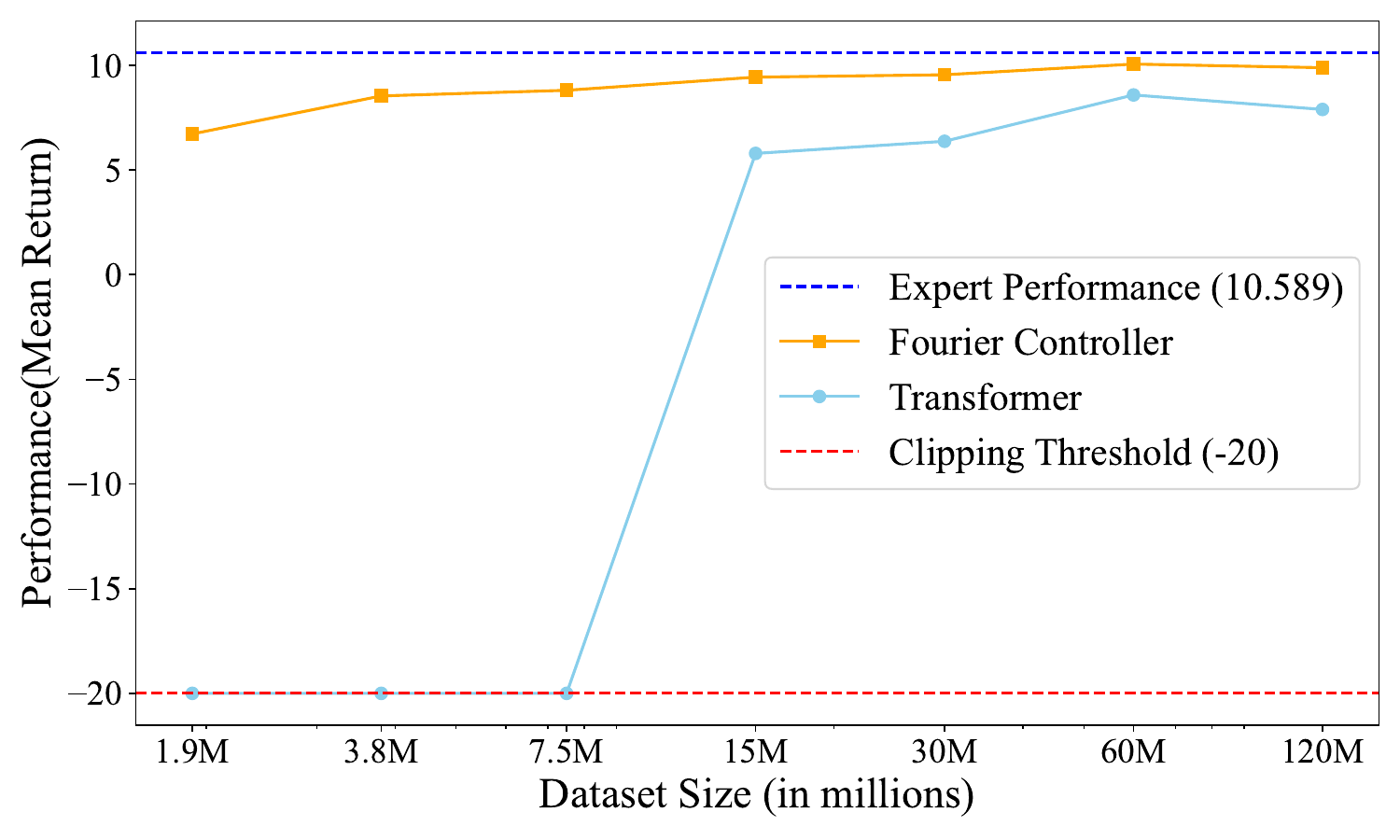}
\caption{The performance of each model on the legged robotics dataset (measured by mean return, averaging the results across 1500*3 trajectories). 
For all future experiments, the 60M-step dataset is utilized as the standard reference.} 
\label{fig:plot_data_size}
\vskip -0.1in
\end{figure}

\paragraph{Evaluation on Simulator.} To assess FCNet's effectiveness on multi-environment robotic datasets, we first perform imitation learning on this dataset. Our comparison focuses on Transformer, as other model architectures like RetNet and MLP have been demonstrated to be less effective in Sec.~\ref{sec:d4rl}. 
We report the performances of FCNet and Transformer across various  dataset sizes in Fig.~\ref{fig:plot_data_size}, where FCNet consistently outperforms Transformer. 
Notably, when the dataset size is comparable to that of D4RL (1M$\sim$ 3M), Transformer's performance is markedly low, whereas FCNet still delivers promising results even with limited data.
This underscores that FCNet, with its inductive bias tailored for robotic control, is more adept at modeling real-world robot trajectories than Transformer.

\begin{figure}[h]
\vskip 0.1in
\centering
\includegraphics[width=0.95\linewidth]{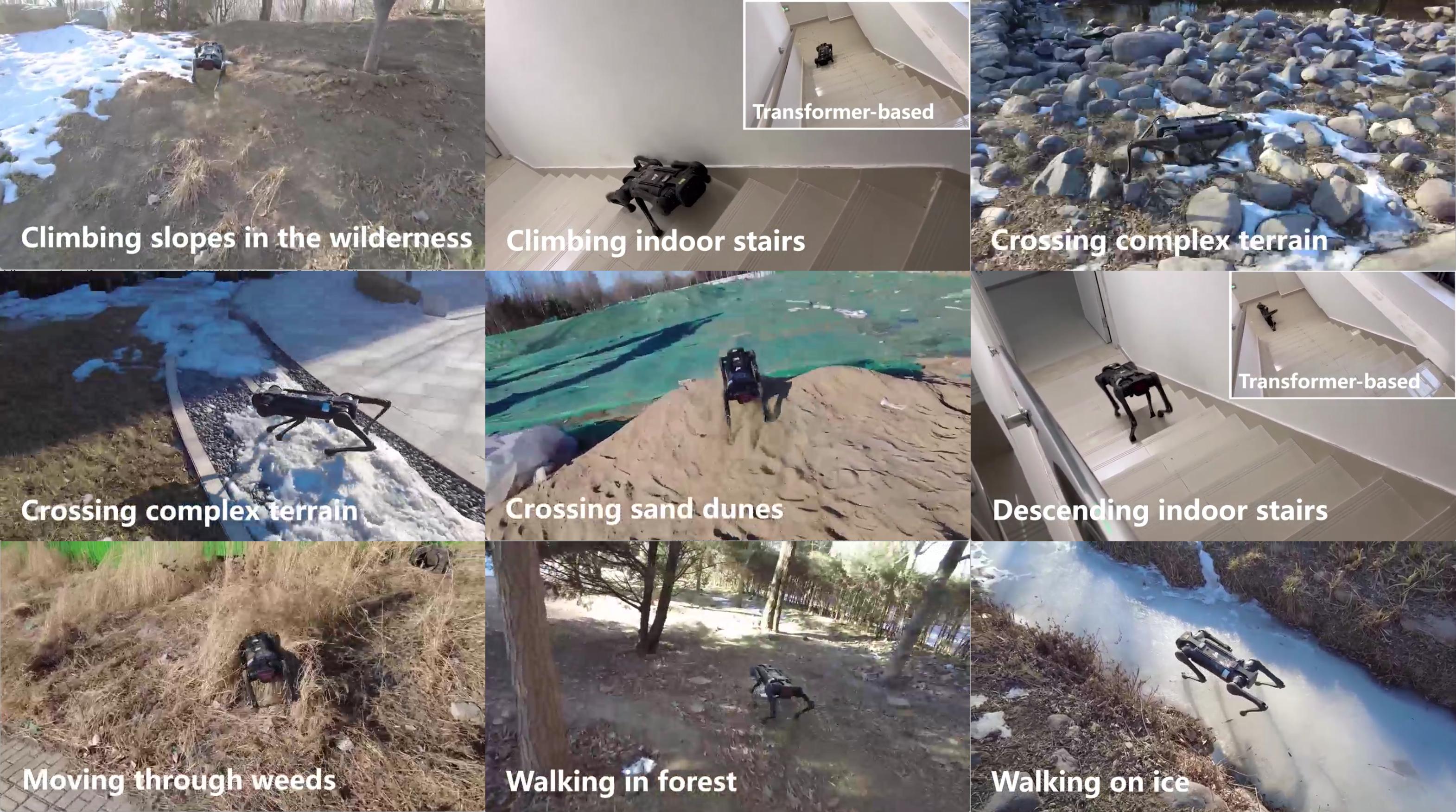}
\caption{Deploying FCNet to real-world legged robots.} 
\label{fig:sim2real}
\vskip -0.1in
\end{figure}

\begin{figure*}[h]
\vskip 0.1in
\centering
\includegraphics[width=0.9\linewidth]{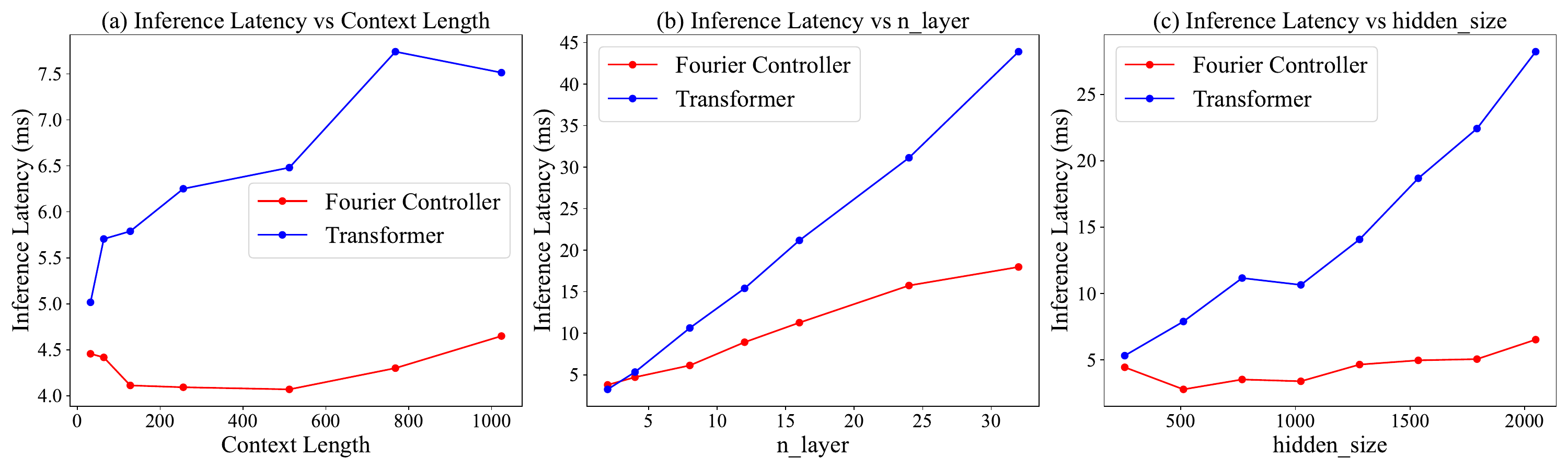}
\caption{The CPU inference latency of FCNet and Transformer under various hyperparameter settings about structure.}
\label{fig:plot_only_arch_speed}
\vskip -0.1in
\end{figure*}
Given the constraints of end-side device performance and the real-time control requirements for robots, it's crucial that the inference latency of an embodied model is kept minimal when deployed in real-world scenarios (e.g., $<20ms$ for quadruped robot). 
\paragraph{Real-World Applications.} 
We further deploy FCNet on real-world robots to evaluate its real-time inference ability and generalizability. 
Training on the 60M-step dataset mentioned in Sec. \ref{para:dataset}, our FCNet shows great adaptability on a variety of terrains not seen in the dataset, such as ice, deep snow, and steep slopes. 
The FCNet-controlled robot moves fluidly, likely due to the filtering of high-frequency noise.
In terms of computational efficiency, FCNet achieves a low inference latency ($\sim$2ms) on less powerful end-side devices, a critical factor for practical deployment.
Conversely, the Transformer model, in our tests, does not perform as well in real-world settings due to issues like lower returns, less smooth output, and longer inference times, leading to less fluid motion and suboptimal performance in indoor scenarios such as stair climbing, turning, and navigating grassy terrains.
More images and videos of these tests are available in the Appendix~\ref{appendix:sim2real} and
supplementary materials.

\subsection{Low Inference Latency} 

To assess FCNet's advantage in inference efficiency, particularly its capability to extract and encode time-varying features via frequency domain interpolation, we conduct a comparative analysis with Transformer under various hyperparameter settings.  
This is done by randomizing input data and  initialized weights. 
We use the empirical value of $m$ in FCNet as $\min\{2.5\log n, \lfloor {n \over 2} \rfloor +1\}$, where $n$ is the context length.
In Fig. \ref{fig:plot_only_arch_speed} (a), as the context length increases, FCNet's inference latency remains relatively stable, whereas Transformer's latency continues to grow significantly, even with a KV cache. This trend can be attributed to Transformer's inherent $\mathcal{O}(n)$ inference complexity.


In Fig. \ref{fig:plot_only_arch_speed}(b) and (c), the inference latencies of FCNet and Transformer are examined as the number of layers and hidden sizes are increased. The results demonstrate that despite the growth in these hyperparameters, the increase in FCNet's inference latency is substantially smaller compared to that of the Transformer. When the model has as many as 32 layers or a hidden size of 2048, FCNet maintains a significant speed advantage, being 3 to 5 times faster than the Transformer. Importantly, even under these conditions, FCNet's inference latency remains below 20ms. This suggests that FCNet, even with an increased number of model parameters, can still meet the requirements for low inference latency. This is a crucial factor for real-time applications in robotics where quick and responsive control is essential.


More comprehensive experiments on ablation study on FCNet and Transformer are included in Appendix~\ref{sec:ablation}, varying the parameter count (controlled by the number of layers and hidden size) and context length.

\section{Conclusion and Limitations} 


In this paper, we present the Fourier Controller Network (FCNet), an innovative model reshaping embodied learning with a focus on frequency domain analysis. FCNet emerges as a distinctive Fourier-based structure, ingeniously incorporating inductive biases for robotic control, and is adeptly designed to minimize time complexity in embodied learning architectures. This design empowers FCNet to extract features with high accuracy, facilitating decision-making in embodied learning through frequency domain interpolation. FCNet excels in training efficiency, showcasing a training complexity of $\mathcal{O}(mn\log n + m^2n)$, and remarkable inference efficiency at $\mathcal{O}(m)$. These developments not only boost the model's effectiveness but also greatly reduce computational requirements. Our results highlight the promise of frequency domain analysis in embodied learning and pave the way for further investigations into pre-training FCNet on extensive robotics datasets, potentially enhancing its generalization abilities significantly.

However, our work has the following limitations: First, we did not validate the scalability of FCNet on a larger embodied dataset. We hope that future scalable methods, such as the attention mechanism, can be adapted in the frequency domain. Second, although we validate the utility of FCNet on complex terrain locomotion and manipulation, we do not consider those scenarios where high-frequency information may be important. Also, we have not yet extended to multimodal inputs. We hope that combining FCNet for online RL training, incorporating multimodal inputs, and scaling FCNet are directions for future exploration.

\section*{Acknowledgements}
This work was supported by the National Key Research and Development Program of China (No. 2020AAA0106302), NSFC Projects (Nos.~ 92248303, 92370124, 62350080, 62276149, U2341228, 62061136001, 62076147), BNRist (BNR2022RC01006), Tsinghua Institute for Guo Qiang, and the High Performance Computing Center, Tsinghua University. J. Zhu was also supported by the XPlorer Prize. We would like to thank Ronghua Hu and Zhuang Zhang for their help in real-world legged robot experiments. We would also like to thank Hao Guo, Muyan Hu, Weilin Zhao, Xinning Zhou, and Huayu Chen for their useful comments. 

\section*{Impact Statement}
This paper presents work whose goal is to advance the field of Machine Learning. The development and implementation of Fourier Controller Networks (FCNet) for real-time decision-making in embodied learning have the potential to significantly enhance the efficiency and effectiveness of robotic systems. 
The application of FCNet in various fields could lead to advancements in automation and intelligent systems, influencing sectors like healthcare, manufacturing, and service industries. 

\bibliography{main}
\bibliographystyle{icml2024}




\newpage
\appendix
\onecolumn

\section{Related Work} 
\label{sec:appA}

\paragraph{RL in Robotic Control.} 
Designing general algorithms and models to control real-world robots, like robotic arms and legged robots, has always been a focus of research.
With the development of Deep Reinforcement Learning, a promising method is to train a general policy via interacting with different dynamics, of which the trained agent can adapt to changes in physical properties, even in some extreme conditions~\cite{ppo,eth2020,eth2022,mit2022, genloco, egocentric, tert, eat, HumanoidTransformer, cmu_xiong2024adaptive}. More recently, pretraining on diverse datasets has shown promise in designing general models. 
DT~\cite{DT} and TT~\cite{TT} treat offline RL as a sequence modeling problem by treating the rewards, states and actions as a sequence so that the Transformer autoregressively generates the next token such as action token. 
Borrowing these ideas, an array of works~\cite{gato, rt1, rt2, vima, robocat, db1, rtx, q_transformer, mobile_aloha, gdm2024autort, gu2023rttrajectory} utilize 
Transformer~\cite{transformer} for pretraining on a larger dataset with multiple decision-making and robotics control tasks, demonstrating strong performance and revealing the generalization that comes with increased model parameters and data size. 
Among them, RT-2~\cite{rt2} uses Internet-scale data to pre-train a Robotics Transformer model with imitation learning methods, and deploys RT-2 on real-world robots for tabletop tasks, which can perform well even in unseen environments. 
However, the large scale of parameters in these models results in high inference latency, which limits their application in real-world robots, thus designing scalable and efficient inference network architecture is significant in this field. 

\paragraph{Efficient Variants of Transformer Architecture.}   Transformer~\cite{transformer} has been heavily used in Natural Language Processing (NLP) tasks since it was proposed. It models the relationship between two by two elements in the context window through the attention mechanism. Due to its high inference time complexity, some works trying to speed up the inference of Transformer such as KV cache, FlashAttention\cite{flashattention, flashattention2} and StreamingLLM~\cite{streamingllm} are proposed to make Transformer have higher speed in inference. However, these works are difficult to achieve theoretically lower complexity while maintaining the same performance. 
\cite{sparse_transformer, Reformer} improves the efficiency of Transformer when dealing with long sequence tasks by improving the attention mechanism, but they do not make improvements in terms of feature extraction capability and inference time complexity. Similarly,~\citeauthor{transformerxl} makes efforts for the efficiency of the Transformer on long sequences of text. 
Meanwhile, many variants of Transformer such as~\cite{linear_transformers, peng2023rwkv, h3, poli2023hyena, retnet, mamba} have been proposed in order to achieve lower time complexity of inference. However, the inductive bias of these model architectures is not appropriate for embodied learning tasks, and it is difficult to achieve strong performance using limited data training. We choose one of the representative State Space Models (SSMs) for experimental validation to demonstrate this. FNet~\cite{fnet}, Fourier Transformer~\cite{fourier_transformer}, and FourierFormer~\cite{fourierformer} explore combining Fourier transforms and Transformer, but do not incorporate feature extraction with low time complexity in the frequency domain, as well as a lack of real-time inference validation on real-world robots. 

\paragraph{Introducing Physical Prior in Machine Learning.}
When dealing with real-world physics, including robotics, it is essential to introduce physical priors into the machine-learning process to decrease data demand and improve models' performance~\citep{karniadakis2021physics,hao2022physics}. For example, PointNet~\citep{qi2017pointnet} and Graph Convolutional Networks (GCN)~\citep{zhang2019graph} make the network architecture permutation-invariant utilizing permutation-invariant operation like summation. And NeuralODE~\citep{chen2018neural} designs an architecture that naturally conforms to ordinary differential equations (ODEs). Fourier neural operator~\citep{li2020fourier}  takes advantage of the symmetry of the differentiation in the time domain and the multiplication in the frequency domain to design a network structure that can efficiently solve partial differential equations (PDEs). Also, FITS~\cite{xu2024fits} discards the high-frequency portion of the time series that has little effect on the time series and fits the time series through a linear layer in the frequency domain. Due to the smoothing properties of most physical phenomena, such structures are also ideal for filtering noise in data. ~\citeauthor{spf, fellows2018fourier} introduce Fourier analysis in reinforcement learning, but do not introduce multi-layer stacked structure for complex pattern learning.

\begin{figure}[h]
\centering
\includegraphics[width=0.5\linewidth]{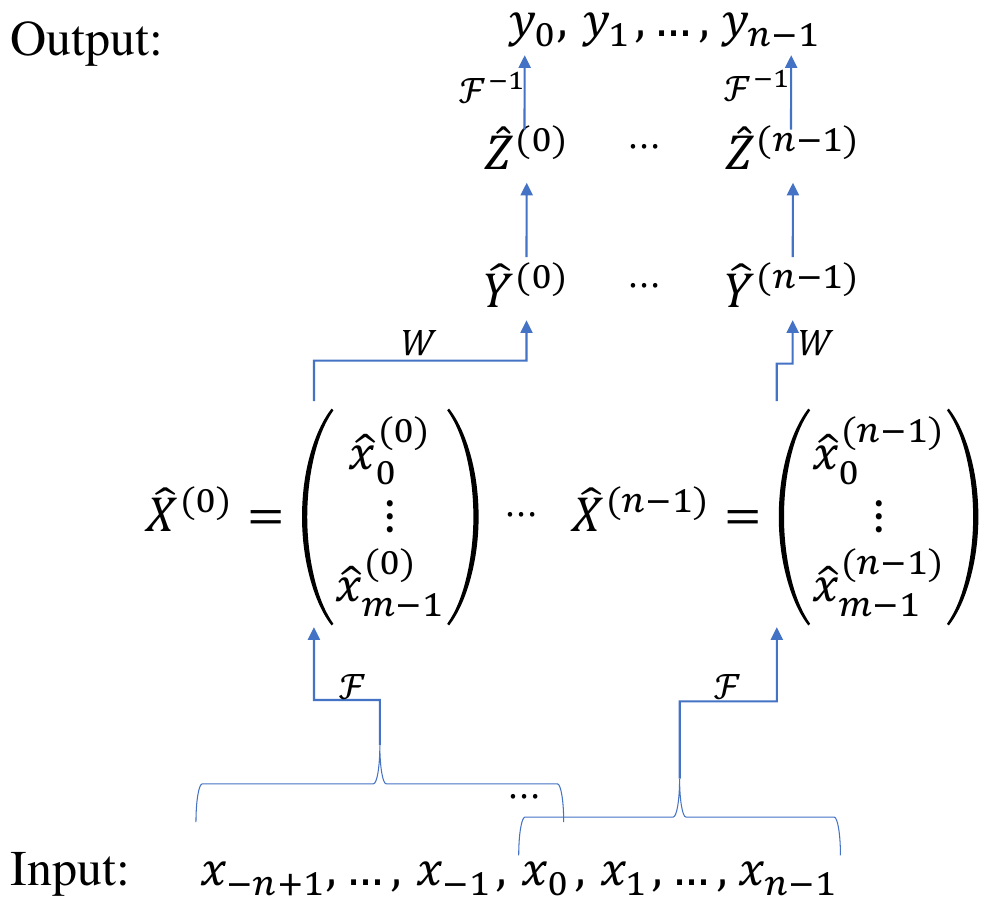}
\caption{CSC block diagram for computing $y_0,y_1,\dots,y_{n-1}$.}
\label{fig:CSC}
\end{figure}

\paragraph{Schematic of the CSC block}
A detailed schematic of the CSC block, illustrating the computation of $y_0,y_1,\dots,y_{n-1}$, 
is shown in Fig. \ref{fig:CSC}.

\section{Ablation Study}
\label{sec:ablation}

\begin{figure}[h]
\centering
\includegraphics[width=0.8\linewidth]{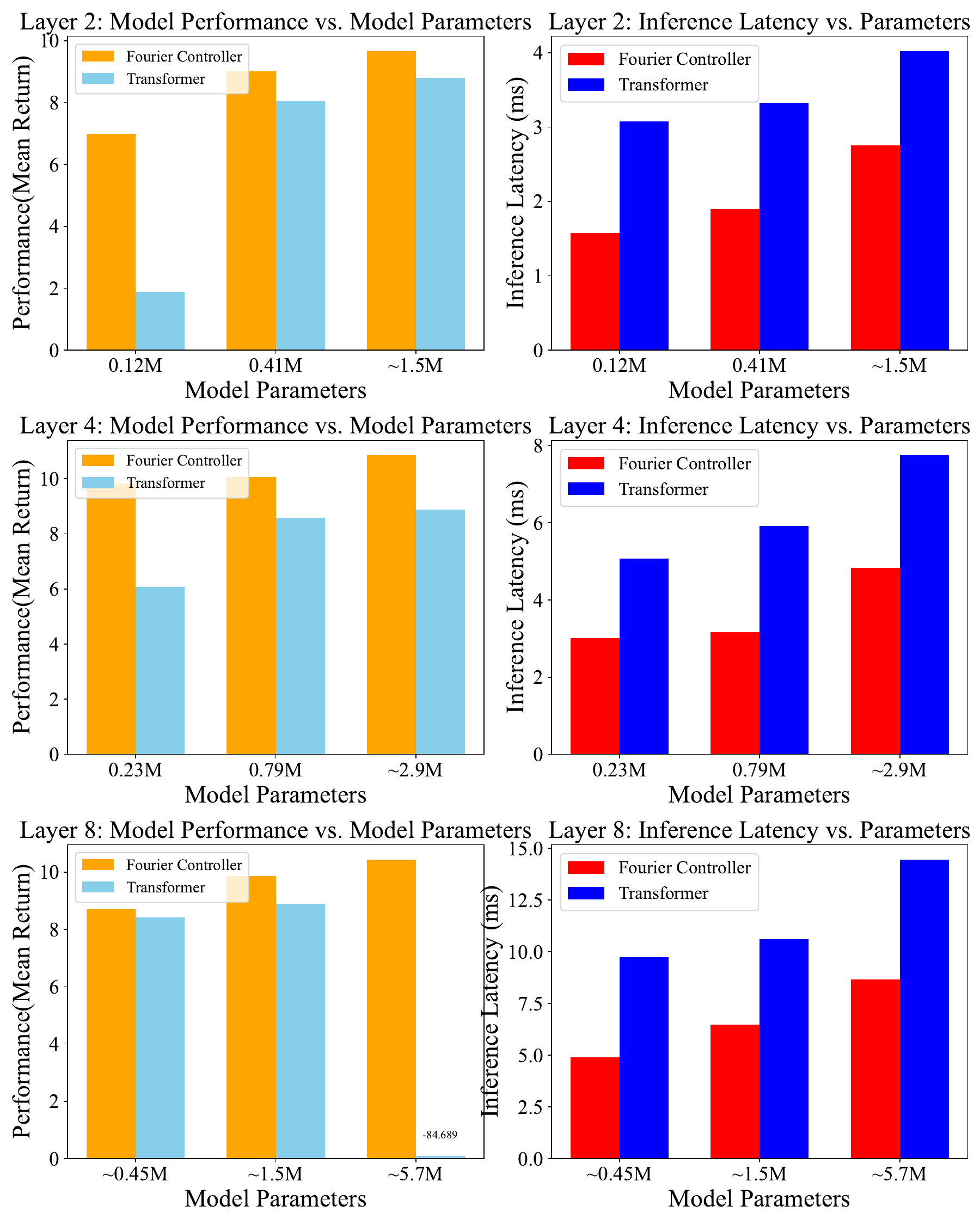}
\caption{The performance and CPU inference latency of each model on the legged robotics dataset (measured by mean return). 
The horizontal axis indicates the model's parameter count, adjustable by varying the number of layers and hidden size in both Transformer and Fourier Controller. Throughout these experiments, hyperparameters like the learning rate remained constant. Notably, in the Transformer experiments with n\_layer=8 and approximately 5.7M model parameters, a high learning rate induces oscillations in the training process.
}
\label{fig:plot_high_speed_inference}
\end{figure}

\paragraph{Varying Model Structures}
In Fig. \ref{fig:plot_high_speed_inference}, FCNet shows consistent superiority in both performance and inference latency over the Transformer across various structures, encompassing different numbers of layers and hidden sizes. Notably, with fewer parameters, FCNet exhibits a robust fitting ability, significantly outperforming Transformer. This implies its effectiveness even in scenarios demanding compact models.  As the model scales up, FCNet maintains greater stability.

\begin{figure*}[h]
\vskip 0.1in
\centering
\includegraphics[width=0.8\linewidth]{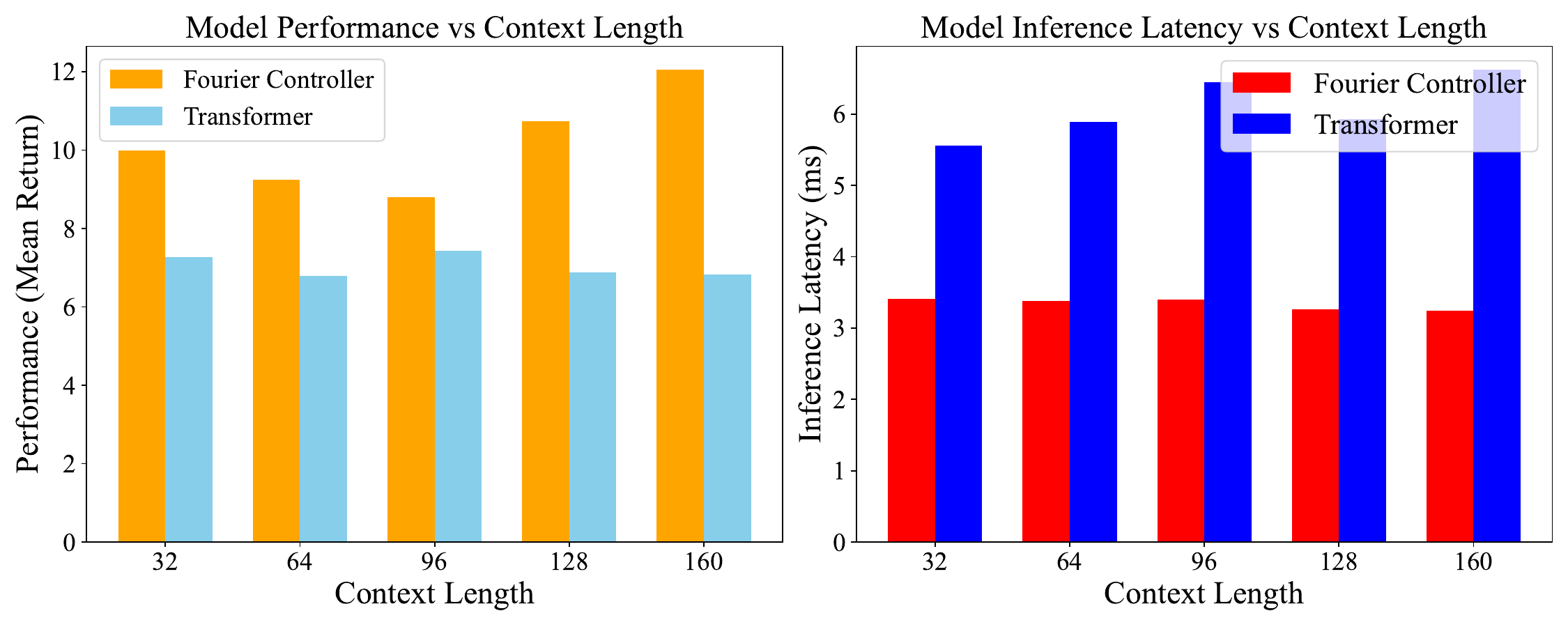}
\caption{The performance and CPU inference latency of each model on the legged robotics dataset (measured by mean return). The horizontal axis indicates the context length of the model. 
This study extends the experiment shown in Fig. \ref{fig:plot_data_size}, where we vary the context length and then evaluate over $1500\times3$ trajectories, averaging both return and inference latency.
We refrained from testing longer sequence lengths because doing so, while keeping the total data volume constant, would decrease the number of trajectories, further leading to data insufficiency.
}
\label{fig:plot_seq_len}
\vskip -0.1in
\end{figure*}

\paragraph{Varying Context Lengths}
Fig. \ref{fig:plot_seq_len} illustrates how FCNet and Transformer perform under various context lengths. With increasing context length, FCNet shows improved performance with relatively stable inference times. In contrast, Transformer does not display the same level of efficiency. Moreover, considering the diverse terrains and historical data in the dataset, context is crucial for policy learning. Experiments reducing Transformer's context length (e.g., to 8) result in completely ineffective policies.

\section{Experimental Details}

\subsection{D4RL}
The hyperparameters of FCNet for D4RL are shown in Table \ref{tbl:gym_hyperparameters}.
\begin{table}[h]
\caption{Hyperparameters of FCNet for OpenAI Gym (D4RL) experiments.}
\vskip 0.15in
\begin{center}
\begin{small}
\begin{tabular}{ll}
\toprule
\textbf{Hyperparameter} & \textbf{Value}  \\
\midrule
Number of layers & $4$  \\ 
Number of modes $m$ & $10$  \\ 
hidden dimension    & $128$ for the last decoder $Q$ \\
                    &   $512$, otherwise \\
Nonlinearity function & GeLU \\
Batch size   & $128$ \\
Context length $K$ & $64$ HalfCheetah, Hopper, Walker \\
Return-to-go conditioning   & $1.15$ medium-expert \\
                            & $1.0$ medium, medium-replay \\
Adam $\beta$ & $(0.9, 0.98)$ \\
Adam $\epsilon$ & $10^{-9}$ \\
LR scheduler & get\_cosine\_schedule\_with\_warmup\\
Learning rate & $5\times 10^{-3}$ \\
Weight decay & $10^{-4}$ \\
Epoch & 50 \\
Learning rate decay & Linear warmup for first $20\%$ training steps \\
\bottomrule
\end{tabular}
\end{small}
\label{tbl:gym_hyperparameters}
\end{center}
\end{table} 


\subsection{Multi-Environment Legged Robot Locomotion Dataset Generation}

The multi-environment legged robot locomotion dataset is an expert dataset, collected in Isaacgym~\cite{isaacgym} by some expert-performance RL policies which have demonstrated strong performance in the real-world legged robot (i.e., Unitree Aliengo). 

It covers various skills (commands from the remote control, e.g., standing, rushing, crawling, squeezing, tilting, and running), and different terrains (e.g., rough terrain, stairs, slopes, and obstacles), following~\citet{eth2022}. The vector of commands issued by the remote control is concated in the state for each time step. 
Terrain-related information is privileged information, i.e., inaccessible, and thus needs to be approximated by historical information to learn an optimal policy. The components of the state include motor velocity, motor position, body angular velocity, and remote control commands. The action in the dataset corresponds to the motor position control of the robot. 

The test version of the dataset contains the aforementioned skills and terrains, with 320,000 trajectories and totaling 60M steps, generated by expert policy. Each trajectory includes the robot's standing and walking on each terrain. 
To maintain both the quantity and diversity of trajectories, we have set an average length of 192 for each trajectory. This length is chosen to comprehensively capture the robot's range of motions, including standing and walking, across various terrains.

\subsection{Multi-Environment Legged Robot Locomotion }
In this experiment, we keep the number of parameters of FCNet and Transformer roughly equal (790k parameters).

The hyperparameters of FCNet for the multi-environment legged robot locomotion dataset are shown in Table \ref{tbl:legged_hyperparameters}.
\begin{table}[h]
\caption{Hyperparameters of FCNet for multi-environment legged robot locomotion experiments.}
\vskip 0.15in
\begin{center}
\begin{small}
\begin{tabular}{ll}
\toprule
\textbf{Hyperparameter} & \textbf{Value}  \\
\midrule
Number of layers & $4$  \\ 
Number of modes $m$ & $10$  \\ 
hidden dimension    & $128$ for the last decoder $Q$ \\
                    &   $256$, otherwise \\
Nonlinearity function & GeLU \\
Batch size   & $1024$ \\
Context length $K$ & $64$ \\
Adam $\beta$ & $(0.9, 0.98)$ \\
Adam $\epsilon$ & $10^{-9}$ \\
LR scheduler & get\_cosine\_schedule\_with\_warmup\\
Learning rate & $5\times 10^{-3}$ \\
Epoch & 50 \\
Learning rate decay & Linear warmup for first $20\%$ training steps \\
\bottomrule
\end{tabular}
\end{small}
\label{tbl:legged_hyperparameters}
\end{center}
\end{table} 

The hyperparameters of Transformer for the multi-environment legged robot locomotion dataset are shown in Table \ref{tbl:trns_legged_hyperparameters}.
\begin{table}[h]
\caption{Hyperparameters of Transformer for multi-environment legged robot locomotion experiments.}
\vskip 0.15in
\begin{center}
\begin{small}
\begin{tabular}{ll}
\toprule
\textbf{Hyperparameter} & \textbf{Value}  \\
\midrule
Number of layers & $4$  \\ 
Number of modes $m$ & $10$  \\ 
hidden dimension    & $128$ for the last decoder $Q$ \\
                    &   $256$, otherwise \\
Nonlinearity function & GeLU \\
Batch size   & $1024$ \\
Context length $K$ & $64$ \\
LR scheduler & get\_cosine\_schedule\_with\_warmup\\
optimizer & Lion \\
Learning rate & $5\times 10^{-3}$ \\
Epoch & 50 \\
Weight decay & $10^{-4}$ \\
Learning rate decay & Linear warmup for first $20\%$ training steps \\
\bottomrule
\end{tabular}
\end{small}
\label{tbl:trns_legged_hyperparameters}
\end{center}
\end{table} 

Other ablation experiments are modified from these hyperparameters. It is also guaranteed that the number of FCNet and Transformer model parameters is close to equal.

A robust sim-to-real FCNet model which we deploy in the real-world legged robot (i.e., Unitree Aliengo for video capture)  is trained from the 60M-step dataset with the metrics shown in Table~\ref{tbl:train_info_sim2real}. We choose $m=10$ and $ n=64$ here ($n$ denotes the context length). In fact, better results might be obtained with larger $n$.
\begin{table}[ht]
\caption{Metrics of sim-to-real FCNet model.}
\vskip 0.15in
\begin{center}
\begin{small}
\begin{tabular}{ll}
\toprule
\textbf{Metrics} & \textbf{Value}  \\
\midrule
Number of model trainable parameters  & $787,852$  \\ 
Loss of training & $0.000408$  \\ 
Loss of test & $0.000420$  \\ 
Mean return in simulator & $10.060/10.589$  \\ 
\bottomrule
\end{tabular}
\end{small}
\label{tbl:train_info_sim2real}
\end{center}
\end{table} 

\subsubsection{Details of observation and action}
We provide detailed information about the observations and actions used in FCNet for the multi-environment legged robotics locomotion task:
\begin{itemize}
    \item Observations:
    \begin{itemize}
        \item  Projected gravity (3-dimensional),
        \item Joint velocity (12-dimensional),
        \item Sine values of joint position (12-dimensional),
        \item Cosine values of joint position (12-dimensional),
        \item Body angular velocity (12-dimensional),
        \item Last action (12-dimensional),
        \item Command (16-dimensional)
    \end{itemize}
    \item Actions
    \begin{itemize}
        \item Expected joint position (12-dimensional).
    \end{itemize}
\end{itemize}

Of note, each '12-dimensional' parameter corresponds to the 12 motors that drive the quadruped robot, including the motors for the hips, thighs, and calves of each of the four legs.

\subsubsection{Sim-to-real gap}
FCNet is pre-trained solely on simulator data before being directly deployed on a real robot to navigate a variety of terrains. This zero-shot transfer leads to robust real-world locomotion, despite the inherent sim-to-real gap.

\paragraph{Sim-to-real gap} The discrepancies between simulators and real-world environments can be seen in areas like state space (differences in robot mechanics), action space (issues like overshooting due to high torque), and transition dynamics (terrain inconsistencies). Additionally, real-world scenarios demand low inference latency to ensure smooth operation - a requirement often not necessary in simulations.

To bridge this gap, we focus on two main areas: data collection and inference.
\paragraph{Data collection} High-quality data is vital. We use reward shaping to guide data generation, avoiding illegal states or actions. The data also needs to encompass a wide range of commands and terrains, achievable through domain randomization, to ensure the model closely aligns with real-world scenarios.
\paragraph{Inference} Post-data collection, engineering efforts such as integrating sensor data into the state and relaying neural network outputs to motor actions become critical. The key to successful sim-to-real transition is the neural network's inference latency. Given FCNet’s low complexity in this regard, we efficiently address the sim-to-real challenge during deployment.

\subsubsection{High-level controller}
The robot's actions, such as moving backward and forward, turning left and right, spinning, standing, crouching, and sprinting, are directed by human operators via remote control. These commands, constituting part of the observation, are interpreted by the policy network, which then translates them into low-level actions executed by the robot.

\subsection{Low Inference Latency}
All our tests for inference latency only target the last action in each context length window, as this ensures that the maximum inference latency of the robot does not exceed a certain hardware threshold when deployed. The testing procedure incorporates three random number seeds, with each seed conducting 10 episodes and subsequently calculating the average of CPU inference latency using \texttt{torch.profiler}, to ensure a small variance.

In the experiments of Fig.~\ref{fig:plot_only_arch_speed}, when we change a hyperparameter, other hyperparameters remain at their default values: \texttt{Context Length}$=64$, \texttt{n\_layer}$=4$, \texttt{hidden\_size}$=256$.

It is worth noting that we tested the model's inference latency on the CPU since many robots lack GPUs, and energy consumption considerations make lightweight deployment crucial.

Our test environment is shown in Table~\ref{tbl:train_info_sim2real2}:
\begin{table}[h]
\caption{Information on the latency measurement environment.}
\vskip 0.15in
\begin{center}
\begin{small}
\begin{tabular}{ll}
\toprule
\textbf{} & \textbf{Value}  \\
\midrule
Warm Up  & $500$ dummy state inputs  \\ 
CPU & Intel(R) Xeon(R) Silver 4210 CPU @ 2.20GHz  \\ 
Memory Configured Clock Speed & 2400MHz \\
OS & Ubuntu 18.04.5 LTS, Linux 4.15.0-135-generic\\
Python & 3.8.18 \\
GCC & 8.4.0 \\
Torch Version & 1.12.1 \\
\bottomrule
\end{tabular}
\end{small}
\label{tbl:train_info_sim2real2}
\end{center}
\end{table} 

Furthermore, in the Ablation Study detailed in Appendix~\ref{sec:ablation}, we consistently measure the inference latency of 8 models simultaneously to ensure environmental consistency across each set of experimental data.

\section{Real-World Applications}
\label{appendix:sim2real}
\begin{figure}[h]
\vskip 0.1in
\centering
\includegraphics[width=0.9\linewidth]{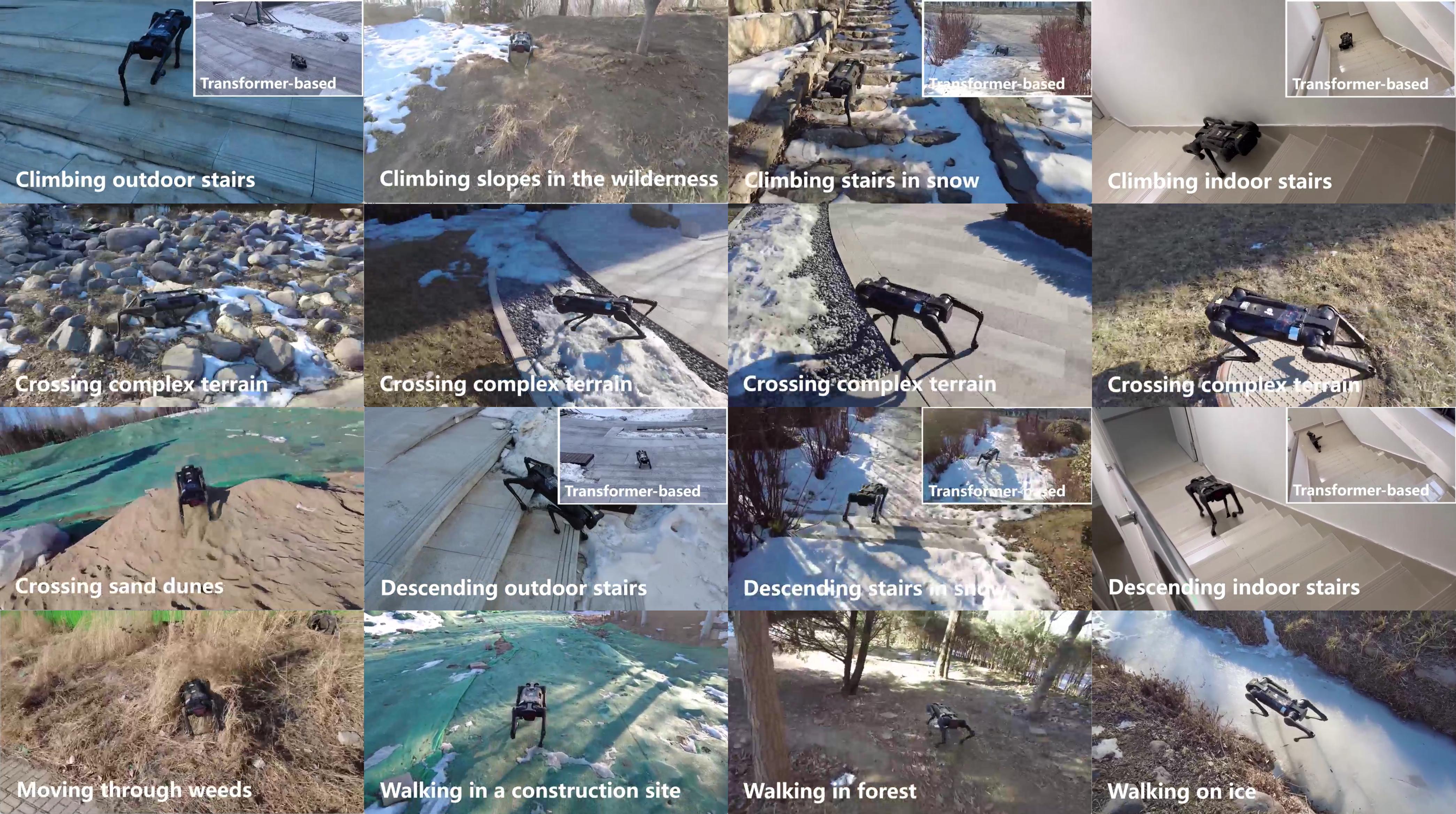}
\caption{Deploying FCNet to real-world legged robots.} 
\label{fig:sim2real16}
\vskip -0.2in
\end{figure}

We provide a more detailed image (i.e., video screen capture) in Fig.~\ref{fig:sim2real16} of real-world robot deployments, as a supplement to Sec.~\ref{sec:legged_exp}. It also includes the comparison of FCNet and Transformer.

In this image, it can be seen that the robot corresponding to Transformer appears to fall in scenarios such as going upstairs indoors and going downstairs indoors. We put a more detailed video with more tests in the supplementary material.




\end{document}